\documentclass[runningheads]{llncs}
\usepackage{amsmath}
\usepackage{graphicx}
\usepackage{tikz-cd}
\usepackage{multicol}
\usepackage{booktabs}

\begin{document}

\title{What If Chain-of-Thought Is Not Enough?}
\author{Marc Jansen}
\institute{Computer Science Institute\\University of Applied Sciences Ruhr West\\ Bottrop, Germany \\ \email{marc.jansen@hs-ruhrwest.de}}
\maketitle

\begin{abstract}
Reasoning capabilities in large language models have evolved from prompting-based methods to internalized model behaviors. In this paper, we investigate whether the internal thinking model of an LLM can be deliberately changed and whether such a change is beneficial. Instead of relying on Chain-of-Thought, we train models to follow Question--Option--Criteria (QOC), a structured decision process that decomposes a task into explicit questions, competing options, and evaluation criteria. We present a three-generation development pipeline that ranges from a basic QOC decision core to richer answer synthesis and statistically enriched QOC reasoning. We then evaluate the resulting models with respect to both internal consistency and output quality. Across the reported experiments, QOC-based fine-tuning improves answer quality relative to non-fine-tuned base models and yields particularly strong gains once models are large enough to execute the underlying QOC computations reliably. The results suggest that the internal thinking model of an LLM is trainable and that structured alternatives to Chain-of-Thought can provide a promising path toward more reliable, interpretable, and task-adapted decision support.
\end{abstract}

\section{Introduction}
This paper investigates a practical next step in the evolution of reasoning-oriented large language models (LLMs): not only eliciting reasoning, but deliberately changing the internal thinking model that a model is trained to follow. Early progress in reasoning quality was largely driven by prompting techniques, most notably Chain-of-Thought (CoT), that encouraged non-thinking base models to externalize intermediate reasoning steps~\cite{wei2022cot,kojima2022zeroshot}. Over time, these externally elicited traces became training signals, enabling models to internalize structured reasoning behavior through rationale-based training and distillation pipelines~\cite{zelikman2022star,hsieh2023distilling}. Our work follows this same development logic, but replaces CoT with a different and more explicitly decision-oriented framework: Question--Option--Criteria (QOC)~\cite{mclean1991qoc}.

The core hypothesis is that QOC offers a better scaffold for decision support tasks because it enforces separation between (i) the decision question, (ii) competing options, and (iii) explicit evaluation criteria. This structure is particularly useful when explainability matters: rather than producing only a final recommendation, the model can expose why certain options dominate and where trade-offs occur, aligning with broader explainability goals in AI decision support~\cite{miller2019explanation,doshivelez2017rigorous}. We therefore aim to improve both decision quality and interpretability by training models to reason in QOC form and then communicate outcomes in user-appropriate detail.

To operationalize this idea, we developed and trained three model generations, each extending the previous one.

In the first generation, the model was trained to produce a QOC thinking structure and to return only the best-performing option to the user. The response layer was intentionally minimal: no extended justification, no narrative explanation, and no additional contextualization beyond the top-ranked option. This generation served as a baseline for testing whether QOC alone can improve internal decision consistency and option ranking quality.

In the second generation, we reused the same QOC thinking structure but changed the response mechanism. Instead of returning only the winning option, we fed the generated QOC structure back into an LLM---specifically, the same model family that produced the structure---and prompted it to interpret the result and provide a more detailed answer. This introduced a two-stage behavior: structured decision formation followed by explanatory rendering. For training, both artifacts were retained: the intermediate QOC reasoning representation and the richer textual response used to communicate the decision.

In the third generation, we further expanded the reasoning structure by incorporating lightweight statistical analysis into the QOC stage. Beyond identifying the best option, the model also computed comparative indicators for near-best options, including dispersion-oriented measures such as standard deviation to reflect stability and sensitivity across criteria. As in generation two, the complete enriched thinking structure was then passed back to the LLM for interpretation into a rich, user-facing answer. Again, both components---the structured reasoning trace and the interpreted response---were included in fine-tuning data for the corresponding model version.

Across these three generations, the development trajectory moves from pure structured selection to structured-and-explained decision support, and finally to statistically informed structured-and-explained decision support. This progression enables a direct evaluation of how much additional explanatory and analytical layers contribute to internal consistency, output quality, and practical interpretability for end users. At the same time, it makes it possible to study a broader question that emerged as central over the course of this paper: whether deeper internal thinking structures are equally useful across model sizes, or whether smaller and larger models benefit from different depths of structured reasoning.

\section{State-of-the-Art}
Recent progress in LLM reasoning started with prompting methods that elicit intermediate steps. Chain-of-Thought (CoT) prompting established the core paradigm for improving multi-step reasoning quality in sufficiently large models~\cite{wei2022cot}. Closely related work showed that even simple zero-shot trigger phrases can induce strong reasoning gains without handcrafted exemplars~\cite{kojima2022zeroshot}. Beyond greedy decoding, self-consistency improved CoT by sampling diverse reasoning paths and selecting stable answers~\cite{wang2022selfconsistency}.

The transition from prompting to model-internal reasoning behavior has also been supported by iterative rationale-based training methods such as STaR, which bootstraps reasoning ability via self-generated and filtered rationales~\cite{zelikman2022star}. In parallel, ReAct demonstrated that combining explicit reasoning traces with environment actions can improve both task performance and interpretability~\cite{yao2023react}. Distillation-based approaches further showed that stepwise reasoning traces can be transferred into smaller models while retaining strong reasoning per\-for\-mance \cite{hsieh2023distilling}. Taken together, these lines of work suggest that reasoning quality does not only depend on scale or prompting, but also on the form of the internal reasoning structure that a model is encouraged or trained to follow.

For the QOC perspective adopted in this paper, the conceptual foundation is design-space analysis with explicit Questions, Options, and Criteria~\cite{mclean1991qoc}. In contrast to CoT-style reasoning, which typically unfolds as a free-form sequential rationale, QOC imposes an explicitly decision-oriented structure by separating the decision question, the available alternatives, and the evaluative criteria used to compare them. This makes QOC especially relevant for settings in which explainability, trade-off transparency, and auditable decision support are important. The present work builds on this perspective by treating QOC not merely as an output format, but as an internal thinking model that can be trained into an LLM and evaluated with respect to both internal consistency and output quality.

\section{The QOC (Question-Option-Criteria) Thinking Model}
In this paper, Question--Option--Criteria (QOC) is used as a structured decision-making process that replaces unconstrained free-form reasoning with an explicit sequence of analytical steps. Instead of asking the model to directly generate an answer, we require it to first represent the problem in terms of a decision question, a set of alternative options, and a set of criteria according to which these options should be evaluated. This makes the reasoning process more transparent and more suitable for supervision, because the intermediate decision structure can be inspected, validated, and reused during fine-tuning.

The QOC process begins with an input task consisting of a question and a supporting context. From this input, the model first formulates the concrete decision question that has to be answered. It then identifies the relevant answer options, that is, the competing alternatives between which the decision will be made. In a next step, the model extracts or formulates the criteria that are relevant for judging these options. These criteria provide the evaluative dimensions of the decision and define what counts as a strong or weak option in the given context.

Once the criteria have been identified, the QOC process assigns an importance weight to each criterion. These weights represent how strongly each criterion should influence the final decision, or in other words, how relevant each criterion is for identifying the final solution or best option. After that, every option is assessed with respect to every criterion. In the formulation used in this paper, this assessment is represented numerically as a score that indicates how well a given option satisfies a given criterion. The central calculation step of QOC then combines structure and quantification: for each option--criterion pair, the criterion weight is multiplied by the corresponding option score. The resulting weighted values are then summed across all criteria to obtain a final overall score for each option.

The best option is the one with the highest total weighted score. In other words, the QOC process turns qualitative judgment into a structured ranking procedure in which the preferred answer is not chosen implicitly, but derived from explicit intermediate calculations. This is important for our work because it allows us to evaluate internal consistency: we can check whether the model identifies reasonable criteria, whether the weighting and scoring steps are performed correctly, whether the option-level products and final sums are correct, and whether the option with the highest final score is in fact selected as the recommendation.

This QOC representation serves different roles across the three model generations studied in the paper. In generation 1, the QOC process is used as the complete decision core: the model produces the QOC structure and returns only the best option as the answer. In generation 2, the same QOC structure is first produced and then passed into a second interpretation step, in which the model translates the structured decision result into a richer natural-language response for the user. In generation 3, the QOC process is further extended with statistical enrichment, so that the final interpretation can draw not only on the weighted QOC scores but also on additional comparative indicators. Across all generations, however, the core QOC procedure remains the same: identify question, options, and criteria; weight criteria; score options; compute weighted totals; and select the best-supported option.

Within the training pipeline, this structured process is not only used at inference time, but also to generate supervision data. Synthetic decision tasks are created, processed through the QOC routine, and transformed into examples for supervised fine-tuning. As a result, the fine-tuned models are encouraged to internalize QOC as their thinking process rather than merely imitate isolated outputs. The purpose of this design is to make decision support more reliable, auditable, and interpretable than in a purely free-form reasoning setup.

\section{Architecture}
This section describes the end-to-end architecture of our QOC-based training approach. Across all three generations, the core system is built around synthetic data generation, structured QOC analysis, and supervised fine-tuning.

The shared pipeline is as follows. First, we generate synthetic decision tasks consisting of a question and a context that are suitable for QOC-style decomposition. Second, each task is processed by a prompt-based QOC agent that executes a fixed analysis routine: identify options and criteria, assign criterion weights, and score how strongly each option satisfies each criterion. Third, the agent derives the best option from the weighted assessment. Finally, generated artifacts are converted into training examples and used to fine-tune the target LLM. This shared architecture is the backbone of all generations; each generation differs only in what additional artifacts are produced and used during answer synthesis.

\subsection{Generation 1 Architecture: QOC-Only Decision Core}
\begin{figure}[htb]
\centering
\includegraphics[width=0.7\textwidth]{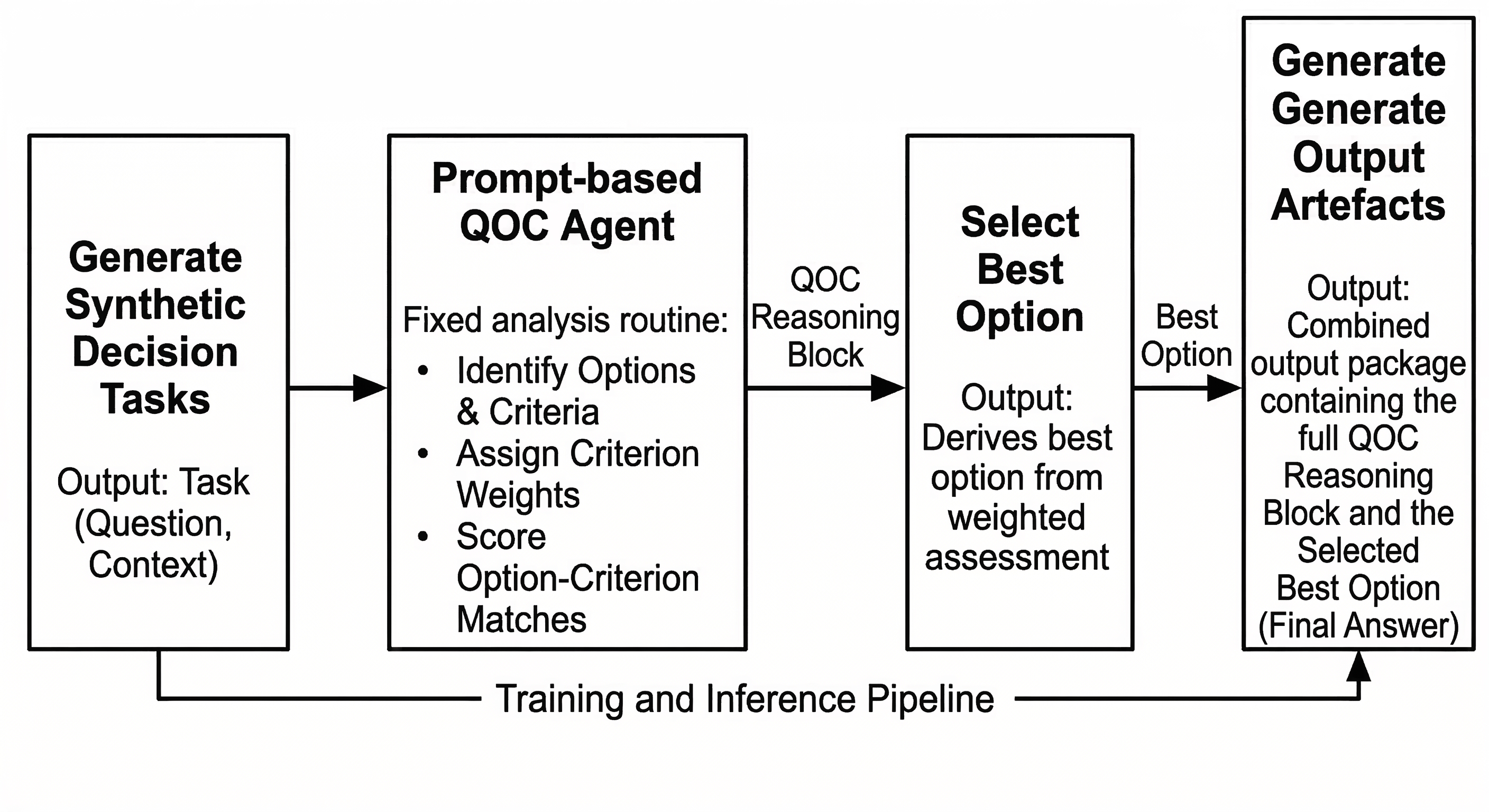}
\caption{Generation 1 pipeline creates synthetic decision tasks, applies structured QOC analysis to identify the best option, and outputs this best option directly.}
\label{fig:arch-gen1}
\end{figure}

Generation 1 uses only the shared core pipeline, as illustrated in Figure~\ref{fig:arch-gen1}. For each synthetic task, the QOC agent produces the reasoning block and selects the best option. The output artifact for training and inference is minimal: the selected best option is returned as the final answer, without an additional interpretation stage for rich natural-language explanation.

\subsection{Generation 2 Architecture: QOC + Interpretation Layer}
\begin{figure}[htb]
\centering
\includegraphics[width=0.7\textwidth]{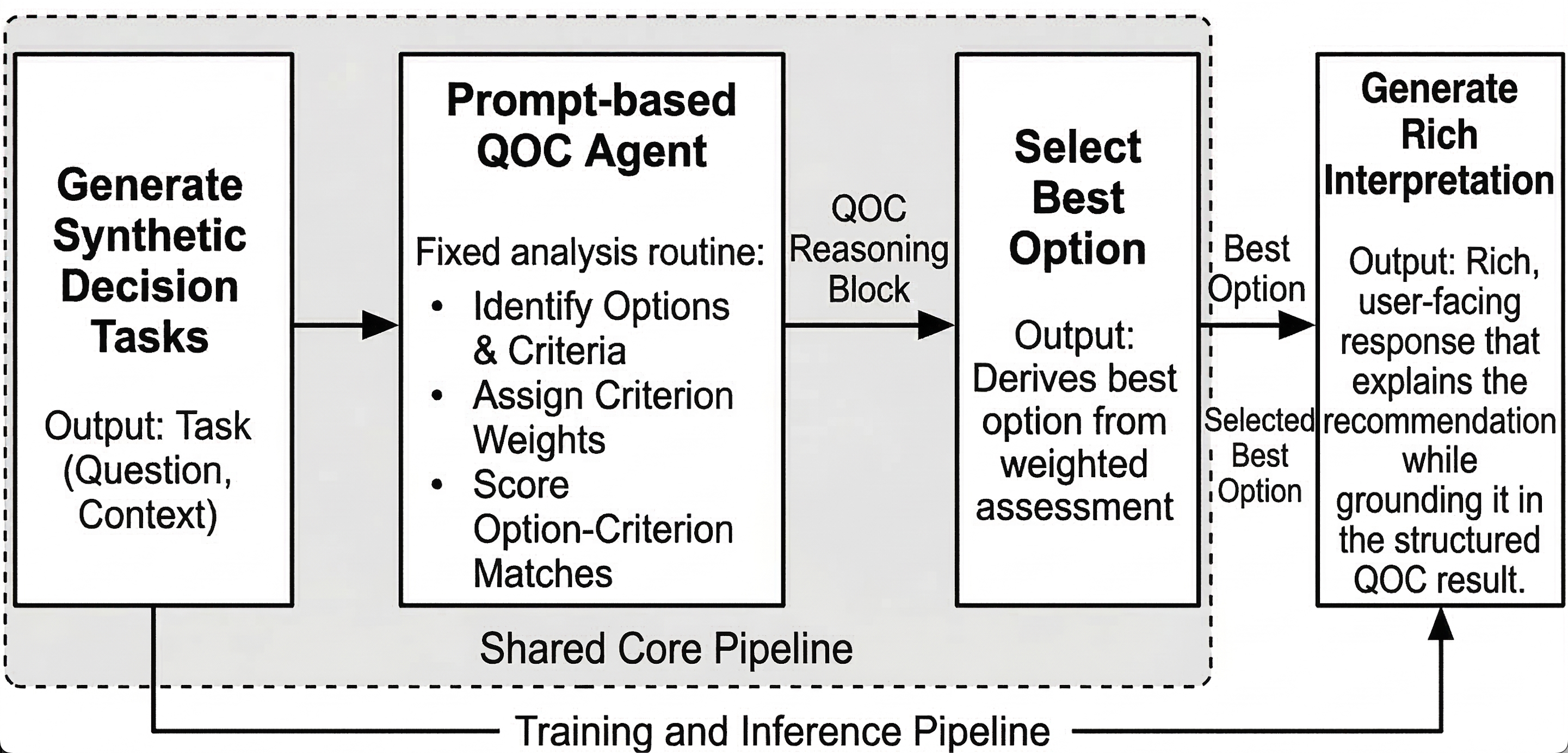}
\caption{Generation 2 architecture integrates a second-stage interpretation module synthesizes a richer natural-language response grounded.}
\label{fig:arch-gen2}
\end{figure}

Generation 2 adds a dedicated interpretation component after best-option selection, as shown in Figure~\ref{fig:arch-gen2}. The QOC thinking block together with the identified best option is passed back to the LLM in a second step. This component generates a richer, user-facing response that explains the recommendation while grounding the explanation in the structured QOC result. Relative to generation 1, the architectural extension is therefore a post-QOC answer-synthesis stage.

\subsection{Generation 3 Architecture: QOC + Statistical Enrichment + Interpretation}

\begin{figure}[htb]
\centering
\includegraphics[width=0.7\textwidth]{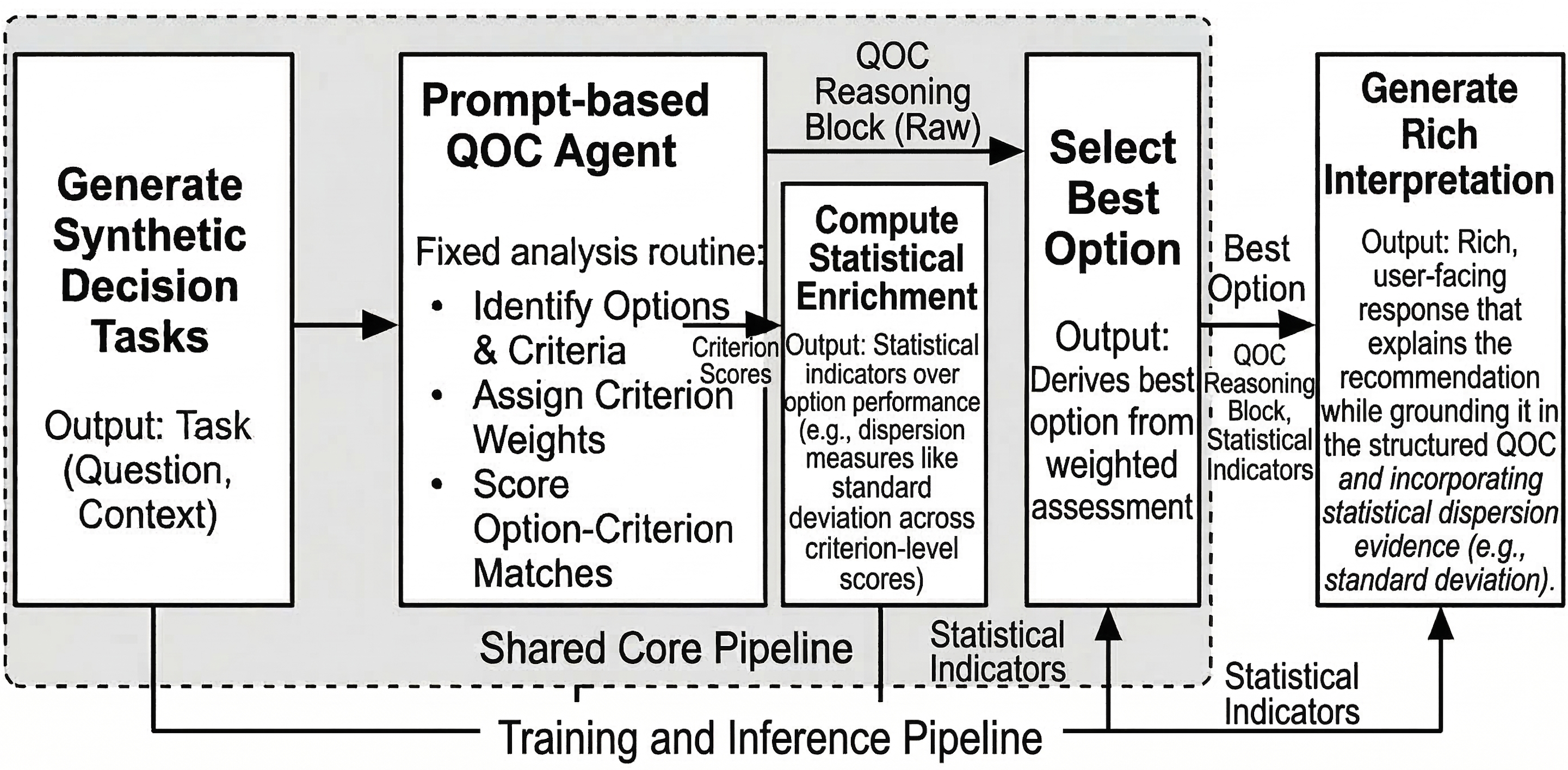}
\caption{Generation 3 pipeline extends the QOC decision core by incorporating statistical enrichment.}
\label{fig:arch-gen3}
\end{figure}

Generation 3 extends generation 2 with an additional statistical enrichment component between QOC analysis and final answer synthesis, as illustrated in Figure~\ref{fig:arch-gen3}. Besides the QOC block and best option, the system computes simple statistical indicators over option performance (e.g., dispersion measures such as standard deviation across criterion-level scores). These statistics are then provided to the LLM together with the QOC result and best option, enabling the final rich response to incorporate both structured decision logic and quantitative comparative evidence.

\section{Implementation}
\begin{figure}[htb]
\centering
\includegraphics[width=0.7\textwidth]{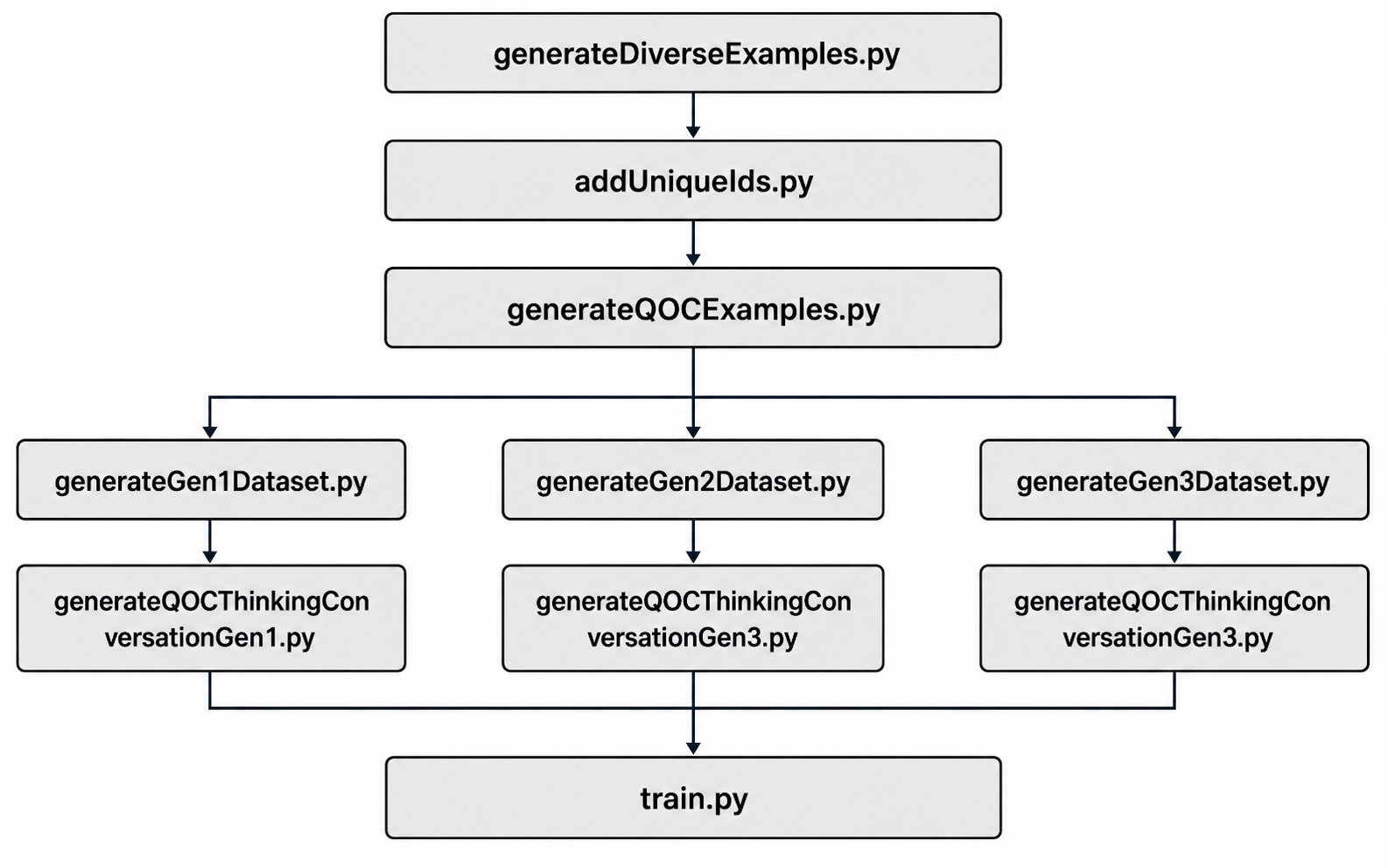}
\caption{Implementation pipeline of the QOC-based training workflow.}
\label{fig:implementation-pipeline}
\end{figure}

This section describes the implementation pipeline that operationalizes the QOC-based training approach introduced in this paper. Figure~\ref{fig:implementation-pipeline} summarizes the end-to-end implementation workflow, from task construction and QOC processing to training data preparation and model fine-tuning. For the generation of the synthetic data, we deliberately rely on the strongest open-weights model available in the respective setup in order to maximize the quality of the generated reasoning traces and supervision signals. In the evaluation setting presented in this paper, this role is fulfilled by Qwen3.5-27B. The pipeline is organized into multiple layers, each of which contributes a distinct function in turning decision tasks into supervised examples for QOC-based model development.

\subsection{Task Generation Layer}
The first layer of the implementation pipeline is responsible for constructing the synthetic decision tasks used throughout the project. A corresponding Python script ensures that these example tasks are drawn from different domains rather than from a single narrow application area. Each task consists not only of a question that should be answered by means of the QOC approach, but also of a synthetic context that provides the background information required for structured decomposition under the QOC framework. This layer therefore defines the raw problem instances on which the remaining parts of the pipeline operate.

\subsection{QOC Processing Layer}
The second layer applies the QOC thinking procedure to each generated task. At this stage, the system uses the relevant decision question, enumerates the available options, derives the criteria for evaluating these options, assigns criterion weights, and scores how well each option satisfies each criterion. The weighted criterion--option values are then aggregated so that the best-supported option can be determined from the resulting totals.

\subsection{Artifact Construction Layer}
Once the QOC analysis has been completed, the implementation pipeline converts the intermediate reasoning outputs into structured artifacts that serve as the data input for the generation of the synthetic training data in the next step. At this stage, the depth of the constructed QOC process depends on the model generation. For generation 1, the constructed representation contains the identified options, the criteria, the criterion weights, and the evaluation of how strongly each option supports the different criteria. The weighted products between option-support values and criterion weights, together with the final sums for each option, are also calculated so that the best option can be identified and reported to the user in the simplest possible form, namely as the final selected option. For generation 2, this process depth is extended by generating a more detailed answer in a second pass through the LLM. This answer is grounded in the QOC analysis and can reflect not only the best option, but also the relative standing of alternative options. In generation 3, the constructed process is extended further with additional statistical measures, for example to identify particularly important criteria or other relevant comparative properties of the option structure. This layer therefore bridges raw reasoning traces and reusable supervision signals for the subsequent generation of synthetic fine-tuning data.

\subsection{Fine-Tuning Data Preparation Layer}
The next layer transforms the generated artifacts into concrete supervised training examples. Based on the QOC data generated in the previous step, the model is additionally asked to analyze the results of the QOC analysis and to produce a final answer derived from this structured evaluation. For generation 1, this simply means adding one sentence that recommends the option with the highest score. For generations 2 and 3, an additional run through the base LLM is performed in order to obtain richer answers and increase transparency. The resulting data is then organized so that the QOC analysis itself is injected into the thinking block of the response, while the generated answer based on this QOC analysis is injected as the final answer of the model. In this way, prompts, intermediate QOC representations, preferred options, and, where applicable, richer natural-language responses are assembled into a format suitable for fine-tuning. The purpose of this layer is to ensure that the target model is not trained only on final answers, but on outputs that reflect the intended QOC reasoning structure.

\subsection{Model Fine-Tuning Layer}
After the training examples have been prepared, they are used to fine-tune the target LLMs. In the context of this paper, this layer corresponds to adapting the different Qwen3.5 base models so that they internalize not only the QOC procedure as part of their reasoning behavior, but also the generation of corresponding final answers derived from this structured analysis. Fine-tuning thus shifts the models from general-purpose base behavior toward a more structured decision-making process grounded in explicit QOC analysis and its answer-level realization.

\section{Evaluation}
This section evaluates the behavior of the generation 2 fine-tuned model, which uses the QOC model as its internal thinking process. Our goal is to assess not only whether the model follows the intended QOC procedure, but also whether this fine-tuning strategy leads to measurable improvements over the underlying non-fine-tuned base models.

We conduct the evaluation along two complementary dimensions. First, we analyze internal consistency, which we define as the extent to which the model correctly performs the calculations required by the QOC framework, including the weighting and scoring steps that determine the final decision. Second, we compare the performance of the fine-tuned models against their corresponding non-fine-tuned base models in order to measure the practical effect of QOC-based fine-tuning on overall model quality.

As the evaluation scenario, we created a benchmark set of 100 example questions, each paired with a corresponding context, drawn from different domains in which the QOC approach is well suited for structured decision-making. We then ran these questions through both the non-fine-tuned base models and the corresponding fine-tuned models across multiple parameter scales. For this evaluation, we used the Qwen3.5 family in all available model sizes: 0.8B, 2B, 4B, 9B, and 27B.

\subsection{Evaluation of the Internal Consistency}

In this subsection, we evaluate the internal consistency of the fine-tuned QOC models. By internal consistency, we mean whether the models correctly perform the calculations required to make decisions according to the QOC framework. In particular, these calculations combine the importance weights assigned to the identified criteria with the scores that represent how strongly each available option satisfies each criterion. Operationally, this first corresponds to multiplying the criterion weights by the respective option-level support values. These weighted values must then be summed to obtain the final score for each option. The recommended answer is the option with the highest overall score. Accordingly, our internal-consistency check does not only examine whether the intermediate calculations and final option sums are correct, but also whether the model identifies the option with the highest sum as the preferred option.

\begin{table}[htb]
\centering
\caption{Internal consistency of QOC-based fine-tuned Qwen3.5 models}
\label{tab:qoc_internal_consistency}
\begin{tabular}{lrrr}
\toprule
\textbf{Model} & \textbf{Correct} & \textbf{Correct Options} & \textbf{Wrong Options} \\
\midrule
Qwen-3.5-0.8B-QOC & 2 & 19 (3.75\%) & 488 (96.25\%) \\
Qwen-3.5-2B-QOC  & 15 & 298 (63.54\%) & 171 (36.46\%) \\
Qwen-3.5-4B-QOC  & 65 & 449 (88.91\%) & 56 (11.09\%) \\
Qwen-3.5-9B-QOC  & 87 & 474 (97.33\%) & 13 (2.67\%) \\
Qwen-3.5-27B-QOC & 95 & 490 (98.99\%) & 5 (1.01\%) \\
\bottomrule
\end{tabular}
\end{table}

In Table~\ref{tab:qoc_internal_consistency}, correct examples indicate benchmark instances for which all QOC-related calculations were performed correctly. Correct and wrong options refer to the number of individual option-level score calculations classified as correct or incorrect, respectively.

The results show a clear and consistent relationship between model size and internal QOC consistency. The smallest model, Qwen-3.5-0.8B-QOC, performs very poorly, solving only 2 benchmark examples fully correctly and producing incorrect option-level calculations in the overwhelming majority of cases. Performance improves substantially at 2B, where the model already produces a majority of correct option-level calculations, but still fails on many complete examples. From 4B onward, the fine-tuned models show strong internal consistency, with the 4B model reaching nearly 89\% correct option-level calculations and the 9B and 27B variants achieving very high consistency. In particular, the 27B model solves 95 of 100 examples fully correctly and reaches almost 99\% correctness at the option-calculation level. Overall, the table indicates that QOC-style fine-tuning becomes substantially more reliable as model capacity increases, with the larger models being able to execute the required weighted scoring and final option selection with high accuracy.

\subsection{Evaluation of the Fine-Tuned vs. the Non-Fine-Tuned Base Models}

In this subsection, we compare the outputs of the fine-tuned QOC models with the outputs of the corresponding non-fine-tuned base models by using an LLM-as-a-Judge evaluation setup~\cite{zheng2023llmjudge}. In this setup, the answer generated by a fine-tuned model is evaluated against the answer generated by a non-fine-tuned model for the same benchmark instance, and a judge model determines which of the two responses is preferable. In our case, the judge model was Google Gemini Flash Lite 3.1. We then measure how often the fine-tuned model response is preferred over the non-fine-tuned response. This comparison is intended to capture practical answer quality rather than internal calculation correctness alone. To assess whether the observed preference counts differed significantly from equal performance, we applied an exact two-sided binomial test to each pairwise comparison in Table~\ref{tab:qoc_judge_comparison}. Under the null hypothesis, the fine-tuned and base model are equally likely to be preferred, corresponding to a preference probability of 0.5. A significant positive deviation indicates that the fine-tuned model was preferred more often than expected under equal performance, whereas a significant negative deviation indicates an advantage of the base model.

\begin{table}[htb]
\centering
\caption{LLM-as-a-Judge preferences for fine-tuned QOC models over non-fine-tuned base models.}
\label{tab:qoc_judge_comparison}
\small
\setlength{\tabcolsep}{4pt}
\begin{tabular}{lccccc}
\toprule
\textbf{Fine-tuned model} & \textbf{27B} & \textbf{9B} & \textbf{4B} & \textbf{2B} & \textbf{0.8B} \\
\midrule
Qwen-3.5-27B-QOC  & 61$^{*}$      & ---          & ---          & ---          & --- \\
Qwen-3.5-9B-QOC   & 53            & 79$^{*}$     & ---          & ---          & --- \\
Qwen-3.5-4B-QOC   & 49            & 81$^{*}$     & 82$^{*}$     & ---          & --- \\
Qwen-3.5-2B-QOC   & 35$^{\dagger}$ & 62$^{*}$    & 66$^{*}$     & 87$^{*}$     & --- \\
Qwen-3.5-0.8B-QOC & 12$^{\dagger}$ & 31$^{\dagger}$ & 26$^{\dagger}$ & 71$^{*}$ & 98$^{*}$ \\
\bottomrule
\end{tabular}
\end{table}

Table~\ref{tab:qoc_judge_comparison} reports how often, across the benchmark set, the answer produced by a fine-tuned QOC model was preferred by the judge over the answer produced by a non-fine-tuned base model. Each cell gives the number of preferences for the fine-tuned model out of 100 pairwise judgments, while the superscript markers indicate whether this observed count deviates significantly from equal performance under an exact two-sided binomial test with a significance threshold of $p = 0.05$. Entries marked with $^{*}$ indicate a significant advantage of the fine-tuned model, whereas entries marked with $^{\dagger}$ indicate a significant advantage of the base model.

The diagonal entries compare each fine-tuned model against its base model of the same size and therefore provide the clearest estimate of the effect of QOC-based fine-tuning. All diagonal entries are marked with $^{*}$, showing that the fine-tuned models outperform their corresponding base models significantly across all tested parameter scales. The improvement is strongest for the smaller models: the 0.8B-QOC model is preferred in 98 of 100 cases over the 0.8B base model, the 2B-QOC model in 87 cases, and the 4B-QOC model in 82 cases. The advantage remains strong at 9B with 79 preferred outputs and is still statistically significant at 27B with 61 preferred outputs. These results indicate that QOC-based fine-tuning yields consistent and statistically robust gains when models are compared against their same-size baselines.

The off-diagonal entries provide a more demanding cross-size comparison. They show how often a fine-tuned model of one size is preferred over a larger non-fine-tuned base model. Here, the significance markers help distinguish robust cross-size gains from comparisons that are compatible with equal performance. For example, the 9B-QOC model is preferred 53 times over the 27B base model, and the 4B-QOC model is preferred 49 times over the 27B base model; neither result is marked as significant, which indicates that these counts do not differ reliably from equal performance. By contrast, the 4B-QOC model is preferred 81 times over the 9B base model and the 2B-QOC model is preferred 66 times over the 4B base model, both with significant positive deviations. At the other end of the spectrum, several entries for the 0.8B-QOC and 2B-QOC models against much larger base models carry the $^{\dagger}$ marker, showing significant negative deviations and thus a clear advantage of the larger base model.

Overall, the table indicates that QOC-based fine-tuning consistently improves the quality of model answers relative to the corresponding non-fine-tuned base models and that these gains are especially pronounced for small and medium model sizes. At the same time, the binomial-test results refine this interpretation by showing where the observed preference counts reflect statistically meaningful deviations from equal performance and where they do not. The gradual reduction of the diagonal advantage from 98 at 0.8B to 61 at 27B suggests that fine-tuning yields the largest relative benefit where the base model is still comparatively weak. For stronger base models, the gains remain positive and statistically significant on the diagonal, but the margin narrows because the underlying model already provides more competitive answers before fine-tuning.

Table~\ref{tab:qoc_bradley_terry} reports Bradley--Terry scores~\cite{bradley1952rank} estimated from the full set of pairwise LLM-as-a-Judge comparisons. Whereas Table~\ref{tab:qoc_judge_comparison} presents the raw preference counts for individual model pairs, the Bradley--Terry analysis aggregates these pairwise outcomes into a single latent strength score for each model. Higher scores indicate that a model was preferred more often across the overall comparison graph. Thus, Table~\ref{tab:qoc_judge_comparison} provides a local view on specific pairwise matchups, while Table~\ref{tab:qoc_bradley_terry} provides a global ranking of all models within a shared preference space.

\begin{table}[htb]
\centering
\caption{Bradley--Terry scores for base and fine-tuned QOC models}
\label{tab:qoc_bradley_terry}
\small
\setlength{\tabcolsep}{4pt}
\begin{tabular}{lcc}
\toprule
\textbf{Model} & \textbf{BT score} & \textbf{95\% CI} \\
\midrule
Qwen-3.5-27B-QOC  & 6.435 & [4.939, 7.932] \\
Qwen-3.5-4B-QOC   & 6.091 & [4.643, 7.539] \\
Qwen-3.5-9B-QOC   & 6.086 & [4.623, 7.549] \\
Qwen-3.5-27B      & 5.988 & [4.546, 7.430] \\
Qwen-3.5-2B-QOC   & 5.282 & [3.844, 6.720] \\
Qwen-3.5-9B       & 4.727 & [3.293, 6.161] \\
Qwen-3.5-4B       & 4.718 & [3.282, 6.153] \\
Qwen-3.5-0.8B-QOC & 3.892 & [2.492, 5.292] \\
Qwen-3.5-2B       & 3.122 & [1.675, 4.569] \\
Qwen-3.5-0.8B     & 0.000 & Reference \\
\bottomrule
\end{tabular}
\end{table}

The Bradley--Terry ranking is broadly consistent with the previously reported pairwise results. The strongest positions are occupied by fine-tuned QOC models, with Qwen-3.5-27B-QOC receiving the highest overall score, followed closely by Qwen-3.5-4B-QOC and Qwen-3.5-9B-QOC. This reflects the earlier observation that fine-tuned models are frequently preferred over their non-fine-tuned counterparts and often remain competitive even in more demanding cross-size comparisons. The ranking also shows that the base Qwen-3.5-27B model remains strong in absolute terms, which aligns with the pairwise table where some smaller fine-tuned models did not achieve significant wins against the strongest base models.

At the lower end of the ranking, the base Qwen-3.5-0.8B model serves as the reference point, while Qwen-3.5-2B and Qwen-3.5-0.8B-QOC occupy comparatively weak positions in the global ordering. This complements the pairwise analysis by showing that, although the 0.8B-QOC model substantially improves over its own base model, it still remains below larger fine-tuned and base models in the overall comparison space. In this sense, the Bradley--Terry table refines the interpretation of the earlier evaluation: the pairwise table identifies where fine-tuning yields significant wins in direct head-to-head matchups, whereas the Bradley--Terry model summarizes these outcomes into a single global ordering that reveals the broader relative strength of all models considered together.

\section{Discussion}
This work provides evidence that the internal thinking model of an LLM can be changed deliberately and that such a change can be beneficial. In our case, the target behavior was not a generic increase in reasoning verbosity, but the adoption of a specific structured decision process: Question--Option--Criteria (QOC). Across the implementation and evaluation results, the central finding is that this structured thinking model can be internalized through fine-tuning and can improve both the internal quality of model reasoning and the externally observed quality of model answers. This is an important result because it suggests that the internal reasoning style of an LLM is not fixed, but can be shaped toward task-specific cognitive structures that are better aligned with particular application domains.

The internal-consistency evaluation makes this particularly clear. The larger fine-tuned models were able to perform the required QOC calculations with high reliability, including criterion weighting, option scoring, weighted aggregation, and final option selection. The pairwise comparison with non-fine-tuned base models further shows that these structured reasoning traces are not merely internally neat, but also translate into better outputs from the perspective of an external judge. Together, these two findings suggest that changing the thinking model of an LLM is not only possible in principle, but can produce practically meaningful benefits when the model is capable of executing the target structure faithfully.

At the same time, the results strongly indicate that the usefulness of increasingly deep QOC-style thinking depends on model size. The smallest models, especially the 0.8B variant, struggle substantially with the internal computational requirements of the QOC framework. Even though fine-tuning still improves their outputs relative to their own base versions, their low internal consistency suggests that these models have limited capacity for reliably executing a fully elaborated QOC process. This points to an important design implication: for small models, simpler forms of structured thinking may be more appropriate, especially thinking processes that do not require complex analytical structures. In the terminology of this paper, such models may be better suited to generation 1, where the core QOC structure is present but the additional computational and interpretive depth remains limited.

The picture changes for larger models. From the middle range upward, and especially from 9B onward, the fine-tuned models show that they can not only benefit from a changed thinking model, but also sustain more profound forms of it. The 9B and 27B models achieve high internal consistency and strong answer quality in pairwise comparison, which suggests that these model sizes are capable of supporting the richer forms of QOC-based reasoning represented by generations 2 and 3. In other words, once the model has sufficient representational and computational capacity, the introduction of a more elaborate internal thinking model appears not to overload the model, but instead to improve its ability to make, justify, and communicate decisions.

A reasonable interpretation of the current results is therefore that all model sizes can benefit from a more structured internal thinking model, but not all model sizes benefit equally from the same depth of structure. Smaller models seem to benefit mainly from the existence of an explicit decision scaffold, while larger models can additionally profit from more detailed intermediate calculations, richer answer synthesis, and possibly even statistically enriched decision representations. This suggests that the choice of thinking model should not be treated as one-size-fits-all, but should instead be adapted to model capacity. The question is not only whether a structured thinking model is useful, but also how much structure a model of a given size can realistically maintain without losing reliability.

These findings also have broader implications for LLM design. Much recent work has emphasized improved reasoning through scaling and through general-purpose reasoning methods such as chain-of-thought prompting and self-improvement supervision~\cite{brown2020gpt3,wei2022cot,zelikman2022star}. Our results instead support a complementary perspective: part of the performance gain may come from selecting and training an appropriate internal reasoning format. In this view, the design of the thinking model becomes a first-class research question. QOC is one example, but the underlying principle is more general. Different tasks may benefit from different internal reasoning models, and the best choice may depend jointly on the task type, the desired form of explainability, and the size of the target model.

There are, however, important limitations to this interpretation. First, the present study evaluates only one family of base models and one structured thinking model. Second, the thresholds suggested here, for example, the idea that smaller models may be better limited to generation 1 while larger models can support generations 2 and 3, should be understood as an informed interpretation of the current data rather than as a universal rule. Third, the evaluation focuses on synthetic tasks and controlled comparisons, which are well suited for methodological analysis but cannot fully capture the complexity of real-world deployment settings.

Even with these limitations, the overall pattern is clear. The paper provides empirical support for the claim that changing the internal thinking model of an LLM is both feasible and often beneficial. It also suggests that the value of deeper thinking structures increases with model capacity. This makes the deliberate design of internal reasoning processes a promising direction for future LLM development and research, especially in domains where decision quality, transparency, and controllable reasoning matter more than unconstrained generative flexibility.


\bibliographystyle{splncs04}
\bibliography{references}

\end{document}